%% file: acl_latex.tex
\pdfoutput=1

\documentclass[11pt]{article}

\usepackage[]{acl}

\usepackage{enumitem}
\usepackage{microtype}
\usepackage{makecell}
\usepackage{inconsolata}
\usepackage{colortbl}

\usepackage{times}
\usepackage{latexsym}
\usepackage{amsmath}

\usepackage{times}
\usepackage{latexsym}
\usepackage[T1]{fontenc}

\usepackage[T1]{fontenc}

\usepackage[utf8]{inputenc}

\usepackage{microtype}

\usepackage{inconsolata}

\usepackage{graphicx}
\usepackage{amsfonts}
\usepackage{amsmath}

\usepackage[most]{tcolorbox}

\usepackage{booktabs}
\usepackage{multirow}
\usepackage{amsmath}
\usepackage{microtype}
\usepackage{graphicx}
\usepackage{amsfonts}
\usepackage{tabularx}
\usepackage{color}
\usepackage{comment}
\usepackage{amsmath,amsfonts,amssymb}
\usepackage{arydshln}
\usepackage{tablefootnote}
\usepackage{xcolor}
\usepackage{subcaption}
\usepackage{float}

\usepackage{bbding}
\usepackage{pifont}
\usepackage{wasysym}
\usepackage{utfsym}
\usepackage{fontawesome}

\usepackage{amsthm}

\title{Revisiting Self-Play Preference Optimization: On the Role of \\ Prompt Difficulty}

\author{
Yao Xiao$^{1,2}$\hspace{1em}
Jung-jae Kim$^{3}$\hspace{1em} 
Roy Ka-wei Lee$^{2}$\hspace{1em}
Lidong Bing$^{1}$\\
$^1$MiroMind AI
$^2$Singapore University of Technology and Design \\
$^3$Institute for Infocomm Research, A*Star, Singapore
}

\begin{document}
\maketitle

\input{latex/abstract}

\input{latex/intro}
\input{latex/background}
\input{latex/exploration}

\input{latex/analysis}
\input{latex/attempt}

\input{latex/relatedwork}

\input{latex/conclusion}
\input{latex/limitation}

\bibliography{anthology,custom}

\appendix

\input{latex/appendix}

\end{document}

%% file: latex/abstract.tex
\begin{abstract}
Self-play preference optimization~\citep{wu2025selfplay} has emerged as a prominent paradigm for aligning large language models (LLMs).
It typically involves a language model to generate on-policy responses for prompts and a reward model (RM) to guide the selection of chosen and rejected responses, which can be further trained with direct preference optimization (DPO).
However, the role of prompts remains underexplored, despite being a core component in this pipeline.
In this work, we investigate how prompts of varying difficulty influence self-play preference optimization.
We use the mean reward of $N$ sampled responses of a prompt as a proxy for its difficulty.
We first find that difficult prompts exhibit substantially inferior self-play optimization performance compared to easy prompts for language models.
Moreover, incorporating difficult prompts into training fails to enhance overall performance and, in fact, leads to slight degradation compared to training on easy prompts alone.
Third, there is a clear upward trend in optimization performance as prompt difficulty decreases.
We also observe that the performance gap between difficult and easy prompts tends to close as the model capacity increases, suggesting that prompt difficulty interacts with the model capacity.
Building on these findings, we explore strategies to mitigate the adversary effect of difficult prompts on final performance.
We demonstrate that only training on a small portion~(30\%) of the easiest prompts improves overall self-play performance on AlpacaEval~2 and Arena-Hard.
We also report failed attempts and lessons learned.

\end{abstract}

%% file: latex/intro.tex
\section{Introduction}
\label{sec:intro}

Large language models (LLMs) have achieved remarkable success in a wide range of natural language processing tasks, but aligning them with human values and preferences remains a challenge~\citep{NEURIPS2020_1457c0d6, wei2022finetuned, bai2022traininghelpfulharmlessassistant, 10.1145/3531146.3533088, bubeck2023sparksartificialgeneralintelligence, grattafiori2024llama3herdmodels, ji-etal-2025-language-models}.
Reinforcement learning from human feedback (RLHF) has become a popular approach to align LLMs with human preferences~\citep{NEURIPS2020_1f89885d, pmlr-v162-ethayarajh22a, NEURIPS2022_b1efde53, pmlr-v235-tang24b, qi2025safety}. 
It involves first training a reward model~\citep{pmlr-v202-gao23h}, which then provides feedback signals to optimize a policy model through reinforcement learning, typically using proximal policy optimization (PPO).  
To further simplify the procedure, \citet{rafailov2023direct} introduced Direct Preference Optimization (DPO), which bypasses the need for reward models when optimizing the policy~\citep{pmlr-v238-gheshlaghi-azar24a, pmlr-v235-ethayarajh24a, meng2024simpo, kim2025preference}.
However, these methods still heavily rely on manually curated pairwise preference data to effectively optimize policy models~\citep{pmlr-v162-ethayarajh22a, pmlr-v235-cui24f, wang-etal-2024-helpsteer, wang2024helpsteer, raghavendra-etal-2025-balancing}.

Recently, DPO-based self-play preference optimization~\citep{wu2025selfplay} has emerged to further enhance the alignment performance of LLMs, which employs standard heuristics or off-the-shelf reward models to select chosen and rejected responses to questions~\footnote{Prompt and question are exchangeable in this paper.} without manual efforts~\citep{tunstall2024zephyr, song2024the, pmlr-v235-chen24j, pang2024iterative, wu2025selfplay, li2025simplemix}.
For example, multiple response candidates can be sampled from policy models and scored with a reward model to construct preference pairs for DPO.
Specifically, the sample with the highest reward is usually selected as the chosen response, while the one with the lowest reward is selected as the rejected response~\citep{meng2024simpo, li2025simplemix}.
Previous work has primarily investigated how reward models and the construction of training pairs contribute to preference optimization and overall alignment performance~\citep{tajwar2024preferencefinetuningllmsleverage, gao2025principled, xiao-etal-2025-finding, qi2025difficultybasedpreferencedataselection},
while the role of prompts has often been overlooked despite being a core component of the pipeline.

In this paper, we fill this gap by focusing on the role of prompts in self-play preference optimization pipeline.
Given an LLM, we first propose using the mean reward of $N$ sampled responses for a prompt as a proxy of its difficulty.
Intuitively, a lower mean reward indicates a higher difficulty for the corresponding prompt.
We can sort prompts by their mean reward to obtain a difficulty ranking, then partition them into subsets of varying difficulty.
Following that, we find that the quartile of prompts with the lowest mean reward (highest difficulty) yields inferior self-play performance than an equal number of prompts in the remaining easier subset.
Furthermore, this quartile of prompts will not lead to performance gains when incorporated into training with the remaining easier prompts.
We then attempt to alleviate the hard prompt issue by proposing three potential approaches.
We find that removing a portion of difficult prompts appropriately will lead to resonable performance gains.
We have following contributions in this work.

First, we use the mean reward of $N$ sampled responses of a prompt as a measure of its difficulty.
Our intuition is that prompts with lower mean rewards are considered more difficult than those with higher mean rewards, allowing us to sort prompts by difficulty.
We find that $10$ samples per prompt suffice to obtain a stable difficulty ranking of the questions.
In addition, we demonstrate that the difficulty of prompts for an LLM can transfer to another LLM to some extent. 
Furthermore, we observe that reward models trained with different loss design and training datasets have similar difficulty assessments, reinforcing the robustness of our metric~(\textbf{Section~\ref{meanreward}}).

Second, we focus on the quartile of prompts with the lowest mean rewards, which corresponds to the most difficult quartile.
We follow the preference pair construction strategy introduced in~\citet{meng2024simpo, li2025simplemix, xiao-etal-2025-finding}, which selects the response of the lowest reward as the rejected and selects the response of the highest reward as the chosen from multiple samples.
We observe that this quartile of prompts tends to yield inferior performance than an equal number of prompts from the remaining set when training through DPO.
In addition, this quartile of prompts will not lead to performance improvement when mixed with the remaining prompts for training.
Furthermore, self-play optimization performance is consistently getting better as prompts become easier~(\textbf{Section~\ref{impact}}).

Third, we attempt to improve the final self-play preference optimization performance of models by mitigating the hard prompt issue: (1) curriculum learning~\citep{Bengio2009CurriculumL} that progressively trains from easy to hard prompts; (2) improving the quality of the chosen response for difficult prompts; and (3) only keeping a small portion of easy prompts (pruning hard ones). 
We find that pruning difficult prompts is simple yet effective, whereas training from easy to hard prompts and improving chosen responses do not translate into final performance gains in our setting~(\textbf{Section~\ref{methods}}).

To conclude, we highlight the overlooked role of prompts in self-play preference optimization in this work. 
We establish mean reward as a practical proxy for prompt difficulty, and show that difficult prompts contribute little to alignment. 
We show the performance difference between hard and easy prompts when optimizing policy models with DPO.
We also share our attempts to mitigate this issue, including the unsuccessful trials.
We encourage future research to revisit the design and utilization of prompts in alignment pipelines, ensuring that self-play preference optimization fully leverages prompts of varying difficulty rather than being hindered by it.

%% file: latex/background.tex
\section{Background}
\label{gen_inst}

\input{figtex/distribution}

\subsection{Direct Preference Optimization}
Different from RLHF, which compresses human preferences into a reward model, DPO~\citep{meng2024simpo} directly aligns language models with human preferences.
DPO is one of the most widely used methods for preference optimization,  
which reformulates the reward function $r$ into a closed-form expression aligned with the optimal policy model.

\begin{equation}
r(x, y) = \beta \log \frac{\pi_\theta(y | x)}{\pi_{\text{ref}}(y | x)} + \beta \log Z(x) \nonumber
\end{equation}

where $\pi_\theta$ denotes the policy model, $\pi_{\text{ref}}$ represents the reference model (usually the supervised fine-tuned checkpoint), and $Z(x)$ is the partition function. 
By embedding this reward formulation into the Bradley-Terry (BT) ranking framework~\citep{19ff28b9-64f9-3656-ba40-08326a05748e}, the probability of preference $p(y_w > y_l | x)$ is calculated as $\sigma(r(x, y_w) - r(x, y_l))$, where $\sigma$ is the sigmoid function. 
Accordingly, DPO circumvents reliance on a reward model by directly leveraging the policy model, which yields the following objective.
\begin{align}
& \mathcal{L}_{\text{DPO}}(\pi_\theta; \pi_{\text{ref}}) = \notag \\
& -\mathbb{E}_{(x, y_w, y_l) \sim \mathcal{D}} \Big[ \log \sigma(r(x, y_w) - r(x, y_l)) \Big] \notag
\end{align}

where $r(x, y) = \beta \log \frac{\pi_\theta(y \mid x)}{\pi_{\text{ref}}(y \mid x)}$.

\subsection{Preference Pair Construction}
\label{conven_pipe}

Given an LLM, a reward function, and a pool of prompts, $n$ candidate responses can be sampled for each prompt from language models. 
The reward function (e.g., a reward model) is then used to score these responses.
In general, responses that receive higher reward scores tend to correspond to higher-quality outputs.

For each prompt, the response with the highest reward is selected as the chosen response, while the response with the lowest reward is selected as the rejected response to form a preference pair.
Previous work has shown that using as few as $5$ samples per prompt is sufficient to achieve significant performance gains~\citep{meng2024simpo, li2025simplemix}. 
In this work, we adopt this pipeline to construct preference pairs of prompts for DPO.

%% file: figtex/distribution.tex
\begin{figure*}[t]
\centering
\begin{subfigure}[t]{0.45\linewidth}
    \centering
    \includegraphics[width=\linewidth, clip=true, trim=0mm 0mm 0mm 0mm]{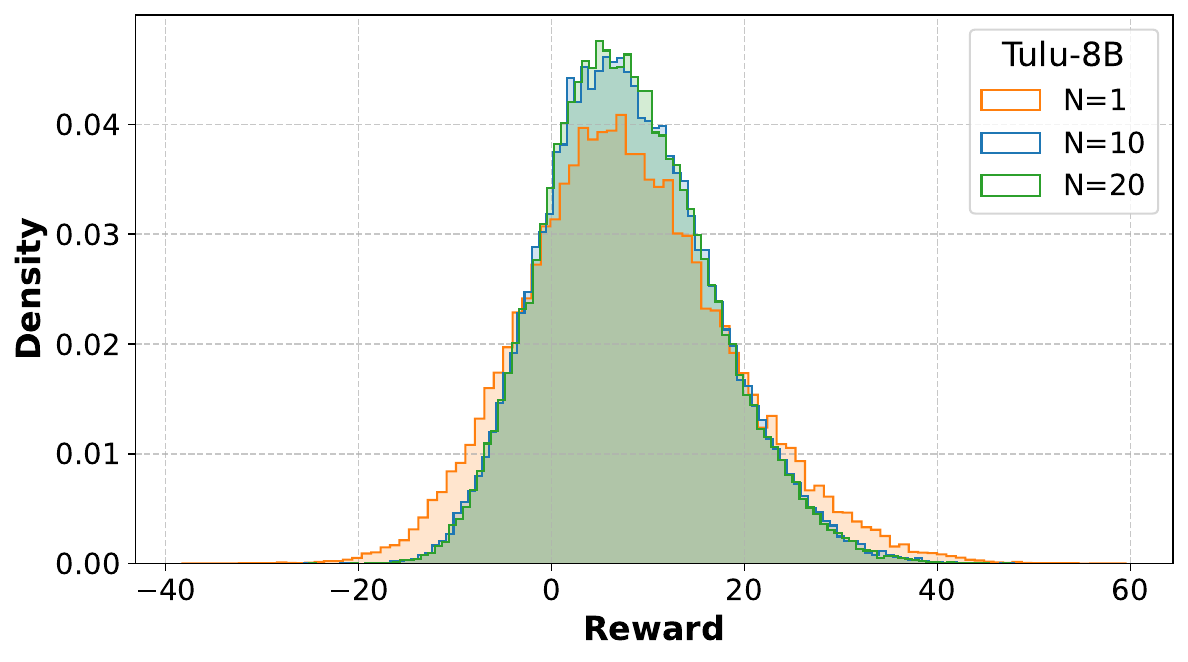}
\end{subfigure}
\hfill
\begin{subfigure}[t]{0.45\linewidth}
    \centering
    \includegraphics[width=\linewidth, clip=true, trim=0mm 0mm 0mm 0mm]{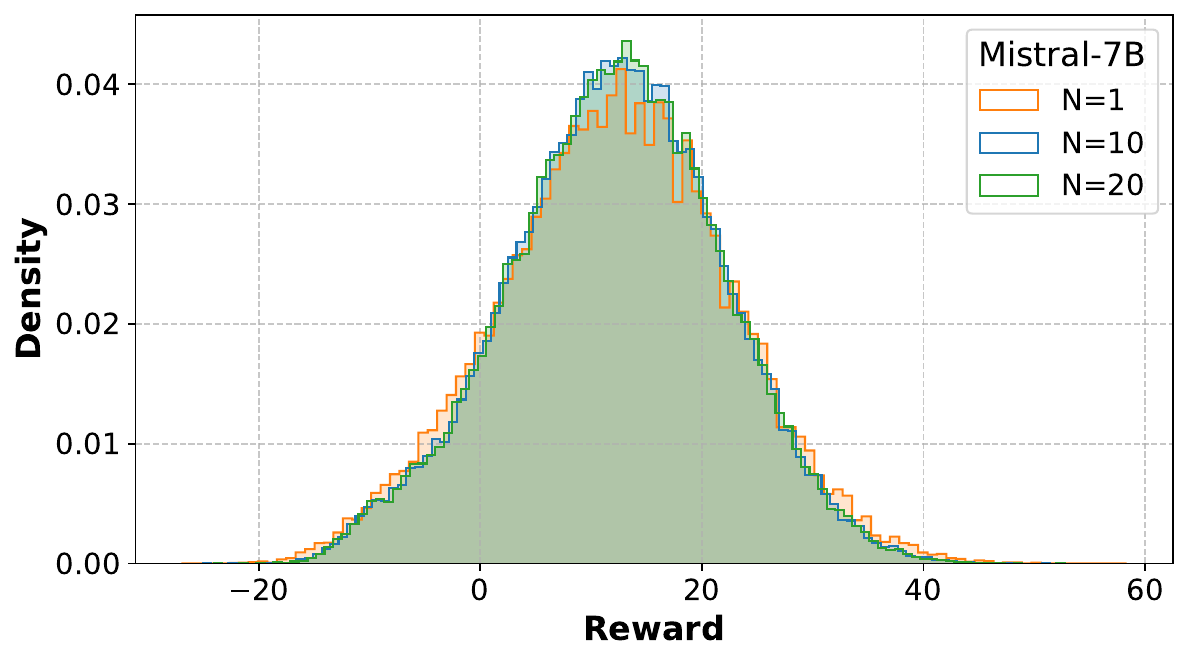}
\end{subfigure}
\caption{We show the mean reward distribution of $N$ sampled responses per prompt on \textsc{Llama-3.1-Tulu-3-8B-SFT} and \textsc{Mistral-7B-Instruct-v0.2} for prompts of UltraFeedback~\citep{pmlr-v235-cui24f}. 
We find $10$ samples per prompt are sufficient to obtain a stable estimate.}
\label{rewarddistribution}
\vspace{-1em}
\end{figure*}

%% file: latex/exploration.tex
\section{Mean Sample Reward as a Proxy of Prompt Difficulty}
\label{meanreward}

The first question in our study of self-play preference optimization is how to quantify the difficulty of a prompt for an LLM. 
In this section, we elaborate on how to measure prompt difficulty based on the mean reward of multiple sampled responses.
In addition, we statistically demonstrate that our difficulty ranking is transferable to some extent between LLMs (RMs).

\subsection{Definition of Prompt Difficulty}
Given a set of questions $\mathcal{D} = \{P_i\}_{i=1}^k$, we sample $N$ candidate responses per prompt from the policy model $\pi_\theta$ and score them with a reward model $r$.
The reward of $N$ candidate samples of the $i$-th prompt is $\left\{r_{ij}\right\}_{j=1}^N$.
The mean of these $N$ reward values is then used as a proxy for the difficulty of the prompt:
\[
D(P_i) = \frac{1}{N} \sum_{j=1}^N r_{ij}.
\]

Our intuition is straightforward: prompts with lower mean rewards are generally harder for LLMs, since they consistently elicit lower-quality responses across multiple samples, whereas prompts with higher mean rewards are easier.
This allows us to rank or partition prompts into subsets according to their estimated difficulty score.

\paragraph{Experimental Details.}
For policy models, we employ \textsc{Llama-3.1-Tulu-3-8B-SFT}~\citep{lambert2024tulu3} and \textsc{Mistral-7B-Instruct-v0.2}~\footnote{For brevity, we may refer to them as Tulu and Mistral in the rest of this paper.}.
To compute rewards, we leverage the publicly available reward model \textsc{Skywork-Reward-Llama-3.1-8B-v0.2}~\citep{liu2024skywork}.
Our implementation is based on vLLM~\citep{kwon2023efficient} for efficient inference with a temperature of $0.8$ and a maximal generation length of $2048$.
Our prompts are from UltraFeedback~\citep{pmlr-v235-cui24f}, which covers more than 61K prompts of high quality from diverse domains.

\input{tabletex/hard_random}

\paragraph{Observation.}
As shown in Figure~\ref{rewarddistribution}, we present the mean reward distribution of $1$, $10$, and $20$ samples per prompt from UltraFeedback~\citep{pmlr-v235-cui24f}.
Generally, increasing the sampling budget produces more stable and consistent estimates of relative prompt difficulty.
In our experiment, we observe that the mean reward of $10$ samples per prompt can provide a stable estimate, which is very close to the distribution of $20$ samples per prompt~\footnote{We also have a Kolmogorov–Smirnov (KS) test between the reward distribution of $10$ and $20$ samples per prompt. 
It turns out that the p-value is lower than $0.05$, which further supports that $10$ samples per prompt are sufficient here.}. 
Therefore, we use the mean reward of $10$ samples per prompt as a proxy for question difficulty throughout this work.

To validate the metric, we sample 200 questions from each quartile of prompts sorted by descending mean reward, assigning difficulty labels {1, 2, 3, 4} accordingly. 
We then ask \textsc{GPT-4o} to independently annotate the those questions with difficulty levels from 1 to 10.
We additionally map the difficulty predictions of \textsc{GPT-4o} into four equally sized bins, yielding labels {1, 2, 3, 4}.
The two labels agree in more than 85\% of cases, indicating a strong alignment between our difficulty metric and the judgments of \textsc{GPT-4o}.

\subsection{Transferability Across Models}
An additional question is whether prompt difficulty, measured by mean reward, is model-specific or generalizable.
To investigate this, we compare the difficulty rank of the whole prompt set for Tulu and Llama.
We find that the Spearman score between Tulu and Llama is about $0.68$.
And there are more than 9,000 common prompts in the most difficult quartile of prompts for Tulu and Llama.
Our results show that prompts identified as difficult for one model tend to remain difficult for another one, suggesting that relative prompt difficulty is not strictly tied to a single policy model but is transferable across LLMs to some extent.
More exploration of transferability between reward models can be found in Appendix~\ref{trans_rm}.

%% file: tabletex/hard_random.tex
\begin{table*}[t]

\centering
\small
\setlength{\tabcolsep}{6pt}
\begin{tabular}{l *{3}{c} *{3}{c}}
\toprule
\multirow{2}{*}{\textbf{Method}}
& \multicolumn{3}{c}{\textbf{Llama-3.1-Tulu-3-8B}}
& \multicolumn{3}{c}{\textbf{Mistral-7B-Instruct-v0.2}} \\
\cmidrule(lr){2-4} \cmidrule(lr){5-7}
& LC (\%) & WR (\%) & Length & LC (\%) & WR (\%) & Length \\
\midrule
Original Model              & 18.10  & 10.90  & 1115 & 17.63 & 14.68 & 1594 \\
\midrule
Hard Prompts & 24.23 & 18.78 & 1502 & 25.96 & 23.11 & 1718 \\
\midrule
\multicolumn{7}{l}{Easier Prompts} \\
\quad Run 1 & 29.61 & 31.93 & 2101 & 28.81 & 29.07 & 2044 \\
\quad Run 2 & 28.15 & 30.99 & 2115 & 28.32 & 28.53 & 2032 \\
\quad Run 3 & 30.83 & 33.35 & 2121 & 28.73 & 27.32 & 2017 \\
\midrule
Average (Easier Prompts) & 29.53 & 32.09 & 2112 & 28.62 & 28.31 & 2031 \\
\bottomrule
\end{tabular}
\caption{The quartile of hardest prompts (bottom $25\%$) underperforms a equal number of easier prompts sampled from remaining $75\%$ on AlpacaEval 2. 
Last row denotes results averaged over three runs for easier prompts.}
\label{hard_to_random}
\end{table*}

%% file: latex/analysis.tex
\section{Impact of Prompt Difficulty on Self-Play Preference Learning}
\label{impact}

\input{figtex/drop_hard}

In this section, we mainly study the hardest quartile (bottom \(25\%\)) of prompts. 
We observe that, given the same number of prompts, hard prompts underperform easy ones. 
Additionally, excluding this quartile of prompts from the full set results in slight but consistent gains in overall performance.
Furthermore, we observe a monotonic improvement in optimization performance as the prompt difficulty decreases.
In the end, we point out that the performance gap between this hard quartile and easier prompts may be closed if policy models are sufficiently capable.

\paragraph{Experimental Details.}
We first sort prompts in UltraFeedback~\citep{pmlr-v235-cui24f} by the difficulty score introduced in Section~\ref{meanreward}.
We use \textsc{Llama-3.1-Tulu-3-8B-SFT} and \textsc{Mistral-7B-Instruct-v0.2} as policy models to sample responses for preference pair construction and further train with DPO~\citep{rafailov2023direct}.
We employ the publicly available reward model \textsc{Skywork-Reward-Llama-3.1-8B-v0.2} to score responses.
We follow the strategy in Section~\ref{conven_pipe} to construct preference pairs by sampling $5$ responses per prompt for DPO.
We evaluate model performance on AlpacaEval~2~\citep{dubois2023alpacafarm, dubois2024lengthcontrolled}, which is the most widely used benchmark in this field.
AlpacaEval~2 consists of $805$ questions from multiple domains and tasks, which enables a comprehensive assessment of LLMs.
Both length-controlled win rate and vanilla win rate~\footnote{For brevity, we refer to length-controlled win rate as LC and refer to win rate as WR in most tables and figures of this paper.} are reported. 
The decoding temperature is $0.9$ and $0.7$ for Tulu and Mistral during evaluation, respectively.
More details about training can be found in Appendix~\ref{hyper}.

\subsection{Harder Prompts Underperform Easier Prompts}
\label{subsec:hard-underperform}
We examine the hardest quartile (bottom $25\%$) of prompts and compare it to subsets drawn from the remaining easier set.
Specifically, we sample an equal number of prompts from the remaining set with three different seeds to ensure a fair and robust evaluation. 
This approach reduces the variance introduced by random selection and provides a more reliable evaluation.
We also aggregate the results by averaging over the three runs, which allows us to better capture the performance difference between the hardest quartile and other easier prompts.
We show their result training through DPO in Table~\ref{hard_to_random}.
Across both backbone models (Tulu and Mistral) and multiple random realizations, performance on the most difficult $25\%$ of prompts exhibits lower performance than that of random subsets from the remaining pool in both length-controlled win rate and vanilla win rate.
These results highlight an underlying weakness of self-play preference optimization when applied to hard prompts.

\definecolor{cream}{HTML}{F7EED7}
\definecolor{edge}{HTML}{2C7A7B}
\begin{tcolorbox}[
  colback=cream,
  colframe=edge,
  boxrule=1.2pt,
  arc=2pt,
  left=6pt,right=6pt,top=6pt,bottom=6pt
]
\paragraph{Observation.}
The hardest quartile of prompts produces smaller performance gains, limiting effective optimization under DPO.
\end{tcolorbox}

\subsection{Hard Prompts Hurt Final Performance}

To further examine the impact of the most difficult quartile of prompts on final performance, we train models with and without this subset under DPO.
In the case without this quartile of prompts, the performance of two backbone models (Tulu and Mistral) improves on AlpacaEval~2 (Figure~\ref{drop_hard}) despite discarding $25\%$ of the training data, which also saves $25\%$ of the computing costs. 
We can observe a slight but consistent improvement in terms of vanilla win rate and length-controlled win rate, indicating the gains are not attributable to longer responses or verbosity effects.
In practice, this means that naive scaling of self-play data without accounting for prompt difficulty may yield diminishing or even negative returns.
It highlights the trade-off between prompt quantity and prompt difficulty, which is of practical significance.

\definecolor{cream}{HTML}{F7EED7}
\definecolor{edge}{HTML}{2C7A7B}
\begin{tcolorbox}[
  colback=cream,
  colframe=edge,
  boxrule=1.2pt,
  arc=2pt,
  left=6pt,right=6pt,top=6pt,bottom=6pt
]
\paragraph{Observation.}
Incorporating hard prompts into training often hurts final performance.
\end{tcolorbox}

\subsection{An Upward Trend as Prompt Difficulty Decreases}

\input{figtex/trend}

We further divide the full prompt set into four quartiles according to their difficulty scores, ordered from the most challenging to the easiest.
For each quartile, we train an independent DPO model, and the resulting performance trend is presented in Figure~\ref{trend}.
A consistent pattern emerges: models trained on easier prompts outperform those trained on more difficult ones, with performance increasing monotonically from the first to the fourth quartile.
This observation highlights the sensitivity of preference-based optimization to the underlying prompt difficulty distribution and suggests that easier prompts provide better optimization performance.

\definecolor{cream}{HTML}{F7EED7}
\definecolor{edge}{HTML}{2C7A7B}
\begin{tcolorbox}[
  colback=cream,
  colframe=edge,
  boxrule=1.2pt,
  arc=2pt,
  left=6pt,right=6pt,top=6pt,bottom=6pt
]
\paragraph{Observation.}
Self-play preference optimization performance can consistently improve as prompts are getting increasingly easy.
\end{tcolorbox}

\input{tabletex/llama_no_gap}

\subsection{The Gap May Be Closed If LLMs Are Strong Enough}
The impact of prompt difficulty is not uniform across LLMs of varying capacity. 
Although weaker policy models such as \textsc{Tulu-3-8B-SFT} suffer from a clear performance gap between hard and easier prompts (Section~\ref{subsec:hard-underperform}), 
this trend does not hold for a stronger model such as \textsc{Llama-3.1-8B-Instruct}. 
Table~\ref{llama_nogap} shows that \textsc{Llama-3.1-8B-Instruct}, a more capable model, achieves comparable performance on hard and easy prompts, with almost no gap in length-controlled win rate and vanilla win rate. 

This finding suggests that the interplay between model capacity and prompt complexity influences the effect of challenging prompts. 
For weaker models, hard prompts tend to underperform easier prompts when performing self-play preference optimization.
In contrast, stronger models are more robust to prompts of varying difficulty.

\textbf{Although sufficiently capable models may bridge the performance gap on hard prompts of UltraFeedback used as a testbed in this work, the existence of hard prompts in the real world remains unavoidable.} 
This underscores the practical importance of our work, which also motivates us to study prompts for self-play optimization.

\definecolor{cream}{HTML}{F7EED7}
\definecolor{edge}{HTML}{2C7A7B}
\begin{tcolorbox}[
  colback=cream,
  colframe=edge,
  boxrule=1.2pt,
  arc=2pt,
  left=6pt,right=6pt,top=6pt,bottom=6pt
]
\paragraph{Observation.}
The performance gap between difficult and easy prompts may diminish when the capacity of LLMs is strong enough.
\end{tcolorbox}

%% file: figtex/drop_hard.tex
\begin{figure*}[ht]
\centering
\begin{subfigure}[t]{0.42\linewidth}
    \centering
    \includegraphics[width=\linewidth, clip=true, trim=0mm 0mm 0mm 0mm]{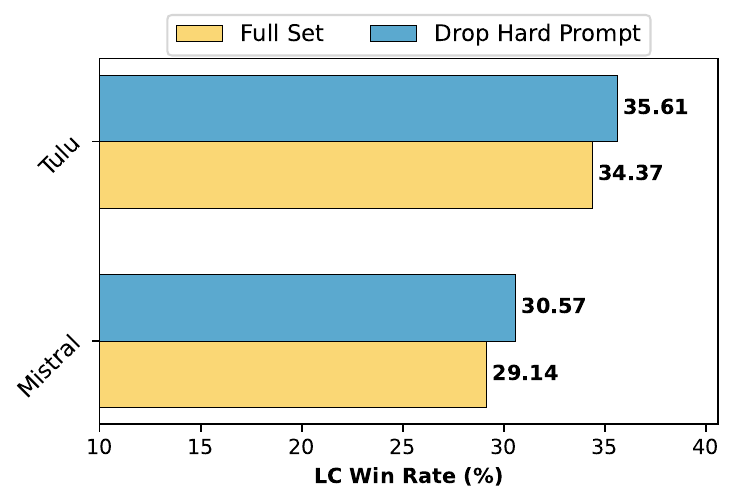}
    \label{}
\end{subfigure}
\hfill
\begin{subfigure}[t]{0.42\linewidth}
    \centering
    \includegraphics[width=\linewidth, clip=true, trim=0mm 0mm 0mm 0mm]{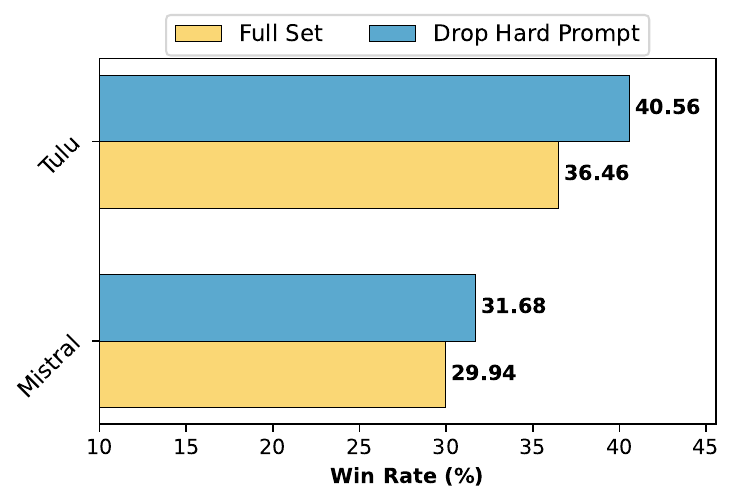}
    \label{}
\end{subfigure}
\vspace{-1em}
\caption{We present the results of dropping the most difficult quartile of prompts and the full set results on AlpacaEval~2.
We can see that incorporating the hardest quartile of prompts into training may hurt the final performance of models.}
\label{drop_hard}
\end{figure*}

%% file: figtex/trend.tex
\begin{figure}[t]
\centering
    \begin{minipage}[c]{\linewidth}
        \centering
        \includegraphics[width=0.49\linewidth]{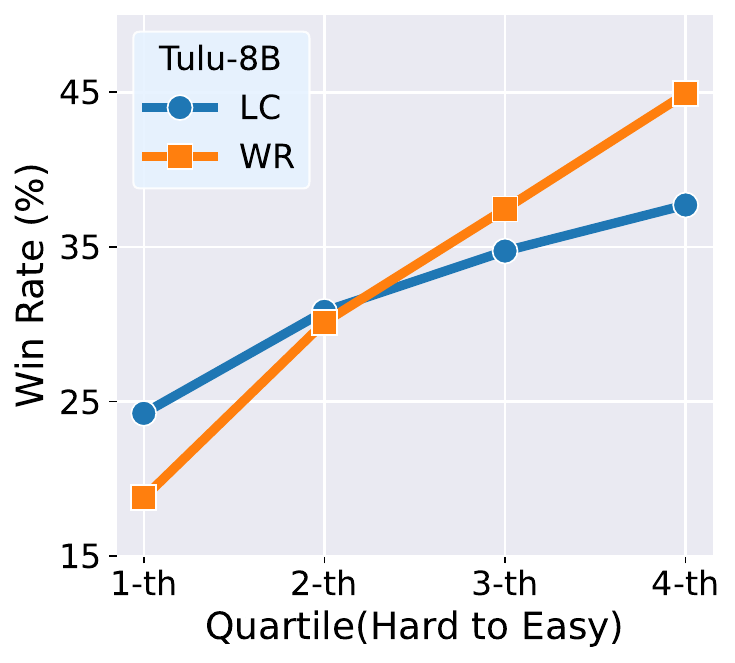}
        \hfill
        \includegraphics[width=0.49\linewidth]{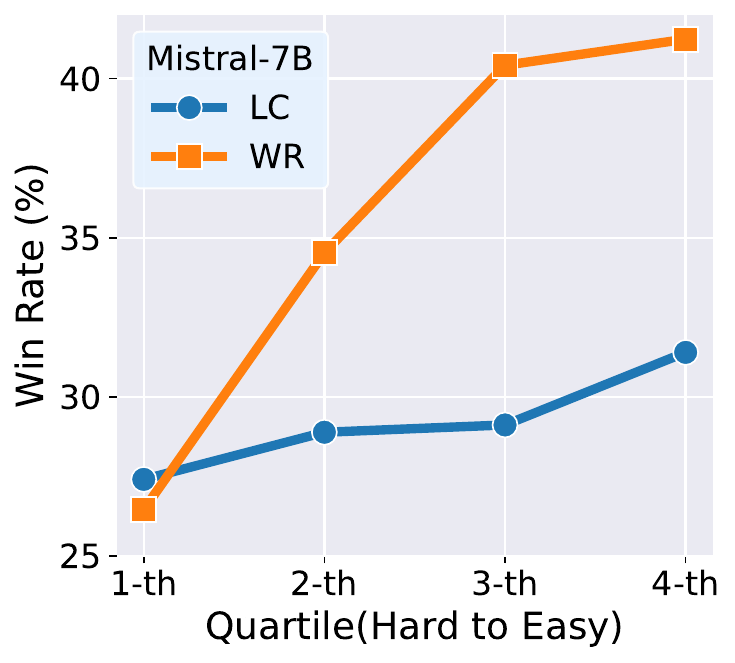}
    \end{minipage}
\caption{Performance improves as prompts are getting easier from first to fourth quartile.}
\vspace{-1em}
\label{trend}
\end{figure}

%% file: tabletex/llama_no_gap.tex
\begin{table}[t]
\centering
\small
\setlength{\tabcolsep}{4pt}
\begin{tabular}{lccc}
\toprule
\textbf{Method} & \textbf{LC(\%)} & \textbf{WR(\%)} & \textbf{Len.} \\
\midrule
Original Model & 23.75 & 24.23 & 1972 \\
\midrule
Hard Prompts & 34.31 & 38.32 & 2059 \\
\midrule
\multicolumn{4}{l}{Easier Prompts} \\
\hspace{1em} Run 1 & 34.59 & 38.32 & 2212 \\
\hspace{1em} Run 2 & 34.45 & 38.62 & 2223 \\
\hspace{1em} Run 3 & 35.25 & 39.19 & 2230 \\
\midrule
Average (Easier Prompts) & 34.76 & 38.71 & 2221 \\
\bottomrule
\end{tabular}
\caption{We find that \textsc{Llama-3.1-8B-Instruct} exhibits no significant performance gap between hard prompts and easier prompts, which suggests that increased model capacity can make model more tolerant to hard prompts to even close the performance gap.}
\label{llama_nogap}
\vspace{-1em}
\end{table}

%% file: latex/attempt.tex
\section{How to Mitigate the Difficult Prompt Issue}
\label{methods}

Our findings suggest that difficult prompts, as identified by their low mean sample rewards, tend to underperform in self-play preference optimization and can even slightly degrade overall performance.
In this section, we investigate three strategies to mitigate the adverse impact of difficult prompts while also documenting unsuccessful attempts to provide a comprehensive account of our study.
We adopt the experimental settings from Section~\ref{impact}, unless otherwise stated.

\input{tabletex/fail_method}
\input{tabletex/main}
\subsection{Unsuccessful Attempts And Implications}
\paragraph{Training from Easy to Hard Prompts.}
Inspired by curriculum learning~\citep{Bengio2009CurriculumL, graves2017automated}, we propose to train models progressively, starting with easy prompts and gradually incorporating more difficult ones.
The difficulty score of the prompts, based on the mean reward of sampled responses, serves as a natural foundation for constructing such a curriculum.
This setup enables us to assess whether curriculum learning can enhance the standard approach of training on the entire prompt set in random order.

As shown in Table~\ref{fail_method}, this curriculum learning paradigm does not improve performance over training on the full dataset in random order for the Tulu model in our setting.

\paragraph{Constructing Better Preference Pairs.}

One hypothesis for the limited contribution of hard prompts is that their chosen responses are more likely to exhibit lower quality, which may hinder self-play preference optimization.
And a language model may struggle to generate high-quality responses within a budget of only $5$ samples per prompt.
Thus, we tried to generate better chosen responses for the most difficult $k$ percent of prompts by increasing the sample budget to $20$ samples per prompt~\citep{xiao-etal-2025-finding}.
By selecting the completion of the maximal reward score in $20$ samples as the chosen response for the most difficult $k$ percent of prompts, we construct less noisy preference pairs for DPO.
In addition, we also tried to sample chosen responses for the most difficult $k$ percent of prompts from a more capable LLM,~\textsc{Llama-3-70B-Instruct}.

The results of these strategies can also be found in Table~\ref{fail_method}.
Both approaches do not produce better performance compared to the results of the full set, whose sample budget is $5$ per prompt.
These findings rule out the possibility that the hard prompt issues arises from low quality chosen responses, and instead suggest that it is fundamentally constrained by the model’s capacity.

\definecolor{cream}{HTML}{F7EED7}
\definecolor{edge}{HTML}{2C7A7B}
\begin{tcolorbox}[
  colback=cream,
  colframe=edge,
  boxrule=1.2pt,
  arc=2pt,
  left=6pt,right=6pt,top=6pt,bottom=6pt
]
\paragraph{Implication.}
Improving the quality of chosen responses for hard prompts does not enhance self-play preference optimization, reinforcing our view that the impact of difficult prompts may be intrinsically constrained by models' capacity.
\end{tcolorbox}

\subsection{A Simple Solution: Train on A Small Portion of Easy Prompts}
Based on our findings in Section~\ref{impact}, which shows that dropping the hardest quartile improves final performance, we examine an adaptive pruning strategy in this section. 
Specifically, we first rank prompts by their difficulty scores and retain the easiest $k$ percent of prompts before constructing preference pairs for DPO.
The value of $k$, which controls the fraction of prompts kept as well as the difficult prompts removed, depends on the prompt difficulty and the capacity of the models. 
In practice, $k$ can be tuned with a benchmark such as AlpacaEval 2.

\paragraph{Experimental Details.}
We mainly follow the experimental setting described in Section~\ref{impact}.
We retain $30$ ($k$=$30$) percent of the easiest prompts for Tulu and Mistral, respectively.
We evaluate model performance on \textbf{AlpacaEval~2}~\citep{dubois2023alpacafarm, dubois2024lengthcontrolled} and \textbf{Arena-Hard v0.1}~\citep{li2024crowdsourceddatahighqualitybenchmarks}.
More details can be found in Appendix~\ref{hyper}.

In addition to results of the full dataset, we also include the results of retaining $30$ percent of prompts for Tulu and Mistral randomly, which are named \emph{Random}.

\paragraph{Results.}
As shown in Table~\ref{tab:main_results}, our method outperforms the complete set and random results on AlpacaEval~2 and Arena-Hard, with significantly less training compute. 
The results support our hypothesis that difficult prompts do not help the preference optimization process, limiting effective learning.
In contrast, removing a controlled portion of the hardest prompts not only improves performance, but also reduces training cost.
This suggests that pruning strategies based on prompt difficulty can serve as a simple yet effective approach to enhance the efficiency of self-play preference optimization pipelines. 
We find consistent results on more reward models, which can be found in Appendix~\ref{more_rm}.

\input{figtex/k_tulu}

\paragraph{Keeping Varying Portion of Easiest Prompts.}
To further investigate the sensitivity of self-play optimization to the proportion of easiest prompts kept, we vary $k$ from $10$ to $50$ and report the results in Figure~\ref{k_tulu}. 
We find that the model performance increases steadily as $k$ grows from $10$ to $30$, with the peak performance observed at $k$=$30$. 
Beyond this point, incorporating more prompts begins to degrade performance.
This trend suggests that incorporating more and more difficult prompts may ultimately hurt optimization. 
In general, these results reinforce the conclusion that prompt difficulty serves as an effective criterion for adaptive pruning in preference optimization.
More experiments can be found in Appendix~\ref{mistral_k_analysis}.

%% file: tabletex/fail_method.tex
\begin{table}[t]
    \centering
    \small
    \setlength{\tabcolsep}{4pt}
    \begin{tabular}{l *{3}{c}}
    \toprule
    \textbf{Method} & \textbf{LC (\%)} & \textbf{WR (\%)} & \textbf{Len.} \\
    \midrule
     Full Set & 34.37 & 	36.46 & 2101 \\
    \midrule
    Easy$\rightarrow$Hard & 33.78 &	34.84 & 2093 \\
    \midrule
    \multicolumn{4}{l}{Chosen in 20 Samples} \\
    \quad $k$=20(\%) & 33.64	& 36.09	& 2129 \\
    \quad $k$=40(\%) & 33.17	& 34.47	& 2071 \\
    \midrule
    \multicolumn{4}{l}{Chosen from Llama-3-70B} \\
    \quad $k$=20(\%) & 34.41  & 38.05 &	2238 \\
    \quad $k$=40(\%) & 26.08  & 32.96   & 2659 \\
    \bottomrule
    \end{tabular}
    \caption{We present our attempts that fail to improve the performance on AlpacaEval~2 on Tulu.
    They are
    (1) training from easy to hard prompts, (2) increasing the number of samples to select the chosen response, (3) sampling chosen responses from more capable models.
    }
    \label{fail_method}
    \vspace{-1em}
\end{table}

%% file: tabletex/main.tex
\begin{table*}[t]
\centering
\begin{tabular}{l *{3}{c} *{3}{c}}
\toprule
\multirow{3}{*}{\textbf{Method}} 
& \multicolumn{3}{c}{\textbf{Llama-3.1-Tulu-3-8B}} 
& \multicolumn{3}{c}{\textbf{Mistral-7B-Instruct-v0.2}} \\
\cmidrule(lr){2-4} \cmidrule(lr){5-7}
& \multicolumn{2}{c}{AlpacaEval 2} & Arena-Hard 
& \multicolumn{2}{c}{AlpacaEval 2} & Arena-Hard \\
\cmidrule(lr){2-3} \cmidrule(lr){4-4} \cmidrule(lr){5-6} \cmidrule(lr){7-7}
& LC (\%) & WR (\%) & WR (\%)
& LC (\%) & WR (\%) & WR (\%) \\
\midrule
Full Dataset & 34.37 & 36.46 & 38.8 & 29.24 & 29.94 & 21.2 \\
Random   & 33.48 & 35.45  & 34.1 & 27.88 & 38.01 & 20.2 \\
\midrule
Ours     & \textbf{38.89} & \textbf{45.78} & \textbf{40.1}  & \textbf{31.43} & \textbf{39.01} & \textbf{23.1} \\
\bottomrule
\end{tabular}
\caption{We report evaluation results on \textsc{AlpacaEval~2} and \textsc{Arena-Hard v0.1}. For our method, we keep only 30\% easiest prompts (lowest difficulty score) on Llama-3.1-Tulu-3-8B and Mistral-7B-Instruct-v0.2 respectively.  \emph{Random} means that we keep 30\% prompts randomly from \emph{full dataset}.}
\label{tab:main_results}
\end{table*}

%% file: figtex/k_tulu.tex
\begin{figure}
    \centering
    \includegraphics[width=0.40\textwidth, clip=true]{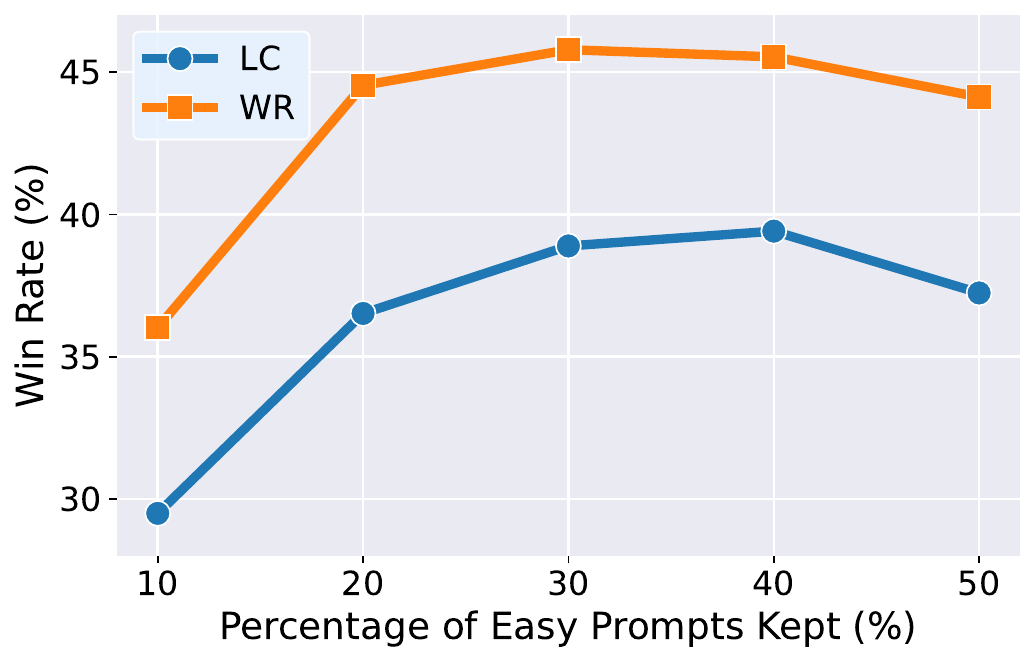}
    \caption{We present the performance of model on AlpacaEval~2 as we change $k$ from $10$ to $50$ percent on Tulu. Performance first improves and then degrades as we retain more and more harder prompts.
    Performance reaches its peak when we train on only $30$ percent of easy prompts (without 70\% hard prompts).}
    \label{k_tulu}
\end{figure}

%% file: latex/relatedwork.tex
\section{Related Work}
\paragraph{Training Sample Selection.}
Training sample selection is a long–standing lever for both generalization and computational efficiency~\citep{NIPS2010_e57c6b95, zhang2025the, deng2025less}. 
For instance, LIMA~\citep{NEURIPS2023_ac662d74} shows that a carefully curated small instruction set can deliver surprisingly effective alignment, reinforcing a “less is more’’ perspective~\citep{muennighoff2025s1simpletesttimescaling}.
In preference learning, sample selection has operated chiefly at the \emph{response} level, e.g., RAFT-style~\citep{dong2023raft} filtering and statistical rejection that keep high-quality chosen responses and prune noisy pairs, or mixing rules for off- vs. on-policy data. Still, recent work sharpens the principle to account for \emph{example difficulty vs. model capacity}. 
Specifically, \citet{gao2025principled} shows that overly difficult preference examples can hinder alignment, which filters such items and yields gains in instruction following~\citep{dubois2023alpacafarm}.

\paragraph{Reinforcement Learning from Human Feedback.} 
RLHF is a leading paradigm for aligning large language models with human preferences in natural language generation~\citep{NEURIPS2022_b1efde53, touvron2023llama2openfoundation}.  
RLHF has been adapted to achieve goals such as reducing toxicity, improving safety, and reasoning~\citep{zhao2023calibrating, qi2024safetyalignmentjusttokens, wu2023finegrained, dai2024safe, yu2024metamath}. 
Nevertheless, it can suffer from training instability and operational complexity inherent to reinforcement learning, as well as its multistage design, which may introduce biases and encourage verbose outputs.
DPO~\citep{rafailov2023direct} and its variants~\citep{meng2024simpo, pmlr-v235-ethayarajh24a, han2024fpogeneralizingpreferenceoptimization} were proposed to mitigate these issues by directly fitting the policy model to pairwise preference data, thereby removing the explicit reward-modeling phase and simplifying optimization. 
As more capable reward models have become publicly available~\citep{jiang-etal-2023-llm, wang-etal-2024-interpretable, wang2024arithmetic, liu2024skyworkrewardbagtricksreward}, a common practice~\citep{dong2023raft, liu2024statistical, meng2024simpo, li2025simplemix} is to use them to score and select self-generated samples, enabling DPO-based training~\citep{tajwar2024preferencefinetuningllmsleverage, dong2023raft, dong2024selfboostinglargelanguagemodels, agarwal2024onpolicy, chen2025longpo, shirali2025direct}.


%% file: latex/conclusion.tex
\section{Conclusion}
\label{conclusion}

In this work, we investigate the often-overlooked role of prompts in self-play preference optimization.
We introduce the mean reward of multiple sampled responses as a practical proxy for prompt difficulty, showing that difficult prompts consistently underperform easier ones in self-play preference optimization.
Through systematic analysis, we demonstrate that incorporating these difficult prompts does not improve performance but slightly degrades it.
We further find that stronger models can nearly narrow this gap, underscoring the interaction between model capacity and prompt difficulty.
We also explore several strategies to mitigate the hard prompt issue. 
Although curriculum training and the construction of higher-quality preference pairs failed to yield improvements, a simple strategy of training on only a small fraction of easy prompts proved effective.
In general, our findings suggest that prompt difficulty deserves careful consideration in self-play preference optimization, ensuring that preference optimization can fully leverage available data without being hindered by inherently difficult prompts.

%% file: latex/limitation.tex
\section*{Limitations}
First, we focus on Direct Preference Optimization (DPO) in this work, which is currently the most widely used preference optimization algorithm. 
However, our conclusions are not specific to DPO and can be extended to other preference optimization methods, which we leave for future work.
Second, we use the mean reward score as a practical proxy for measuring prompt difficulty. 
This metric may not be perfect, considering its accuracy can be influenced by biases in the training data and loss design of reward models.
But it offers a cheap, scalable, and applicable solution. 
We further validate its effectiveness through LLM-as-Judge comparison and additional empirical results.
Looking forward, we plan to investigate ensembles of reward models drawn from diverse families to reduce bias and further improve the robustness of our difficulty estimates.

%% file: latex/appendix.tex
\section{Appendix}

\subsection{Transferability between Reward Models}
\label{trans_rm}
We also explore the difficulty transferability between different reward models, \textsc{Skywork-Reward-Llama-3.1-8B-v0.2} and \textsc{ArmoRM-Llama3-8B-v0.1}~\citep{wang-etal-2024-interpretable}.
They are training with different loss design and data.
We find that the difficulty scores between them computed with 10 samples per prompt have a Spearman score about 0.72. 
In addition, there are more than 9.4k common prompts in the most difficult quartile between them.
This result shows that prompts identified as difficult for a reward model tend to remain difficult for another one, suggesting that prompt difficulty is transferable across reward models to some extent.

\subsection{Implementation Details}
\label{hyper}

\paragraph{Training Hyperparameter.}
For all of our experiments, we use Trl~\citep{vonwerra2022trl} for training. 
Initially, we performed hyperparameter sweeps for $\beta = \{0.01, 0.05, 0.1, 0.5\}$ and max learning rate 
$= \{3e{-}7, 5e{-}6, 1e{-}6\}$ as initial exploration.
We train \textsc{Llama-3.1-Tulu-3-8B-SFT}
with $\beta = 0.01$ and max learning rate $= 3e{-}7$ and train  \textsc{Mistral-7B-Instruct-v0.2} with $\beta = 0.05$ and max learning rate $= 3e{-}7$
We employ 10\% of steps as training warmup.

\paragraph{Evaluation Hyperparameter.}
We maintain a maximal generation length, 2048, for both models. 
The temperature for \textsc{Mistral-7B-Instruct-v0.2} is 0.7 and is 0.9 for \textsc{Llama-3.1-Tulu-3-8B-SFT} when evaluating AlpacaEval~2.
For Arena-Hard, the maximal length is 4098 for both models and we use default temperature 0.

\subsection{Results on More Reward Models}
\label{more_rm}
In this part, we demonstrate the effectiveness of removing the most difficult $k$ percent of prompts for \textsc{Llama-3.1-Tulu-3-8B-SFT} on more reward models, \textsc{ArmoRM-Llama3-8B-v0.1}~\citep{wang-etal-2024-interpretable} and \textsc{Llama-3.1-8B-Instruct-RM-RB2}~\citep{malik2025rewardbench2advancingreward}.
Specifically, we remove the most difficult 70\% of prompt and evaluate the model after DPO training with AlpacaEval~2.
The results are shown in Table~\ref{rms_results}.
Our conclusions hold across different reward models.

\input{tabletex/rewardmodels}

\subsection{Keeping Varying Portion of Easy Prompts for Mistral}
\label{mistral_k_analysis}

We vary $k$ from 10\% to 40\% and report the results in Figure~\ref{k_mistal}. 
There is an increasing trend as $k$ grows from 10\% to 30\%, with the peak performance observed at $k=30\%$. 
Beyond this point, incorporating more difficult prompts begins to degrade performance.

\input{figtex/k_mistral}

%% file: tabletex/rewardmodels.tex
\begin{table}[ht]
  \centering
  \small
  \setlength{\tabcolsep}{8pt}
  \begin{tabular}{lccc}
    \toprule
    \multirow{2}{*}{\textbf{Method}} & \multicolumn{3}{c}{\textbf{ArmoRM-Llama3-8B-v0.1}} \\
    \cmidrule(lr){2-4}
    & LC & WR & Length \\
    \midrule
    full   & 36.39 & 39.13 & 2313 \\
    random & 35.12 & 34.67 & 1933 \\
    our    & \textbf{39.45} & \textbf{41.34} & 2032 \\
    \midrule
    \multirow{2}{*}{\textbf{Method}} & \multicolumn{3}{c}{\textbf{Llama-3.1-8B-Instruct-RM-RB2}} \\
    \cmidrule(lr){2-4}
    & LC & WR & Length \\
    \midrule
    full   & 37.59	 & 37.83 & 2045 \\
    random & 35.03   & 34.26 &	2044 \\
    our    & \textbf{40.09}	 & \textbf{41.67} &	2096 \\
    \bottomrule
  \end{tabular}
  \caption{We demonstrate the validity of removing hard prompts on more reward models, \textsc{ArmoRM-Llama3-8B-v0.1} and \textsc{Llama-3.1-8B-Instruct-RM-RB2}.}
  \label{rms_results}
\end{table}

%% file: figtex/k_mistral.tex
\begin{figure}[H]
    \centering
    \includegraphics[width=0.48\textwidth, clip=true]{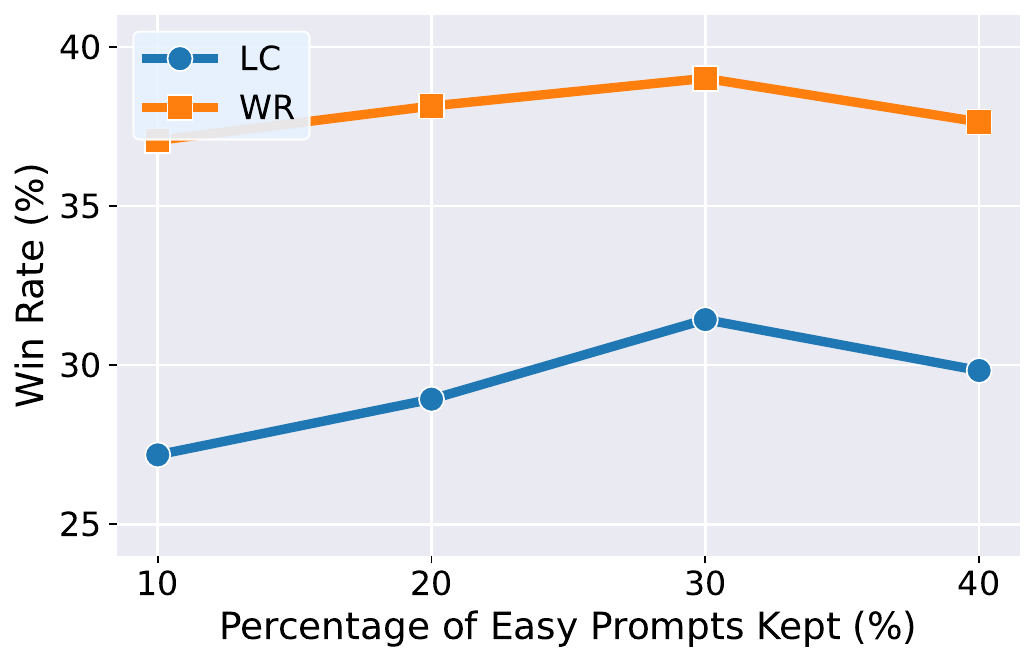}
    \caption{We present the performance of model on AlpacaEval~2 as we change $k$ from 10 to 40 percent on Mistral. 
    Performance first improve and then degrade as we incorporate more and more hard prompts.
    Performance reaches peak when we use only about $30\%$
    percent of easiest prompts.}
    \label{k_mistal}
    \vspace{-0.5em}
\end{figure}